\documentclass[10pt,twocolumn,letterpaper]{article}

\usepackage{cvpr}
\usepackage{times}
\usepackage{epsfig}
\usepackage{graphicx}
\usepackage{amsmath}
\usepackage{amssymb}

\usepackage{color}
\definecolor{blue}{rgb}{0,0,1}
\definecolor{red}{rgb}{1,0,0}
\definecolor{black}{rgb}{0, 0, 0}

\newcommand{\bj}[1]{{\color{black}\textbf{}#1}\normalfont}
\newcommand{\yx}[1]{{\color{black}\textbf{}#1}\normalfont}

\usepackage[breaklinks=true,bookmarks=false]{hyperref}

\cvprfinalcopy 

\def\cvprPaperID{****} 

\begin{document}

\title{Directing DNNs Attention for Facial Attribution Classification using Gradient-weighted Class Activation Mapping}

\author{Xi Yang\textsuperscript{1} \qquad
Bojian Wu\textsuperscript{1,2} \qquad
Issei Sato\textsuperscript{1} \qquad
Takeo Igarashi\textsuperscript{1}\\
\textsuperscript{1}The University of Tokyo \qquad
\textsuperscript{2}SIAT
}

\maketitle

\begin{abstract}

Deep neural networks (DNNs) have a high accuracy on image classification tasks. However, DNNs trained by such dataset with co-occurrence bias may rely on wrong features while making decisions for classification. It will greatly affect the transferability of pre-trained DNNs. In this paper, we propose an interactive method to direct classifiers paying attentions to the regions that are manually specified by the users, in order to mitigate the influence of co-occurrence bias. We test on CelebA dataset, the pre-trained AlexNet is fine-tuned to focus on the specific facial attributes based on the results of Grad-CAM.

\end{abstract}

\section{Introduction}

\begin{figure}[h]
\begin{center}
  \includegraphics[width=0.3\linewidth]{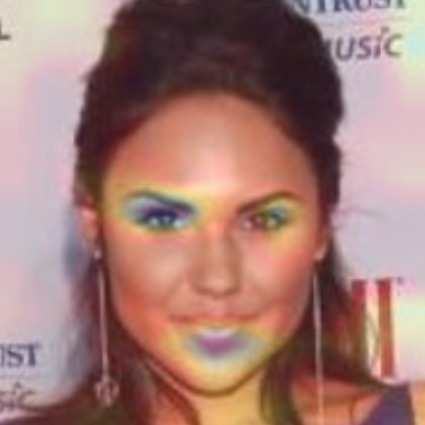}~
  \includegraphics[width=0.3\linewidth]{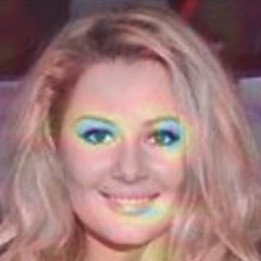}~
  \includegraphics[width=0.3\linewidth]{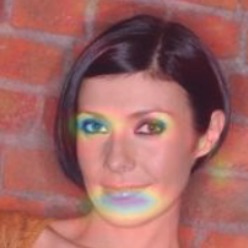}
  \\
  \includegraphics[width=0.3\linewidth]{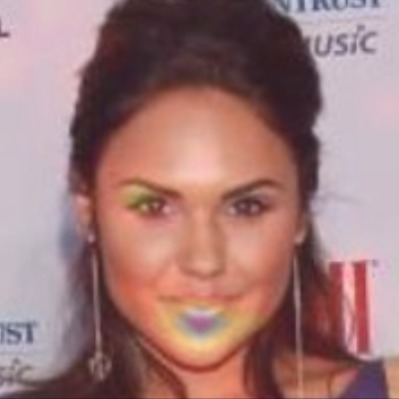}~
  \includegraphics[width=0.3\linewidth]{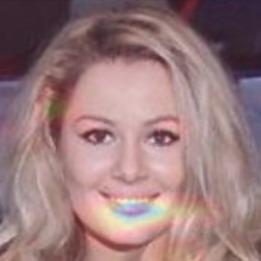}~
  \includegraphics[width=0.3\linewidth]{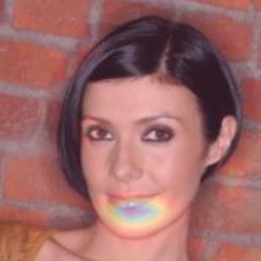}
\end{center}
   \caption{Examples from the CelebA dataset. Grad-CAM shows that a pre-trained DNN `lipstick' attribute focuses on not only the mouth region but also on the eyes, eyebrows, and other regions (first row). Our fine-tuned DNN focuses on the mouth only (second row). The predictive importance of various facial regions (high to low) is colorized blue to red.}
\label{fig:biased}
\end{figure}

Many datasets feature various biases frequently, such as co-occurrence bias, which is attributed to a lack of negative examples \cite{torralba2011unbiased}. Strong correlations among several featured elements mean that the feature one wishes to extract is often accompanied by other features. A network trained by datasets featuring obvious biases cannot reliably make decisions based only on desired features, as proven by the \textit{the lipstick problem} of \cite{Zhang2018ExaminingCR} shown in Figure \ref{fig:biased}. Although an attention approach may be used to improve DNN performance \cite{fukui2018attention}, biased representations may still be in play.

In Large-scale CelebFaces Attributes (CelebA) dataset \cite{liu2015faceattributes}, for example, the attributes `Wearing Lipstick' and `Heavy Makeup' often occur simultaneously with a high probability. Most people in the images, not only applying lipstick but also putting makeup on other facial parts, will be only labelled by the attribute of `Wearing Lipstick'. Thus, the network recognizes `Wearing Lipstick' usually relying on the makeup of several parts of a face, such as the eyes, eyebrows, and mouth. In this way, a pre-trained network lacks strong transferability from one dataset to another, because of bias in representation. To improve DNN generalization, it is important to focus on the correct extracted features facilitating classification; accuracy is not everything.

The biased representations can not be simply eliminated from a pre-trained network; it is difficult to disentangle extracted features. Apart from directly modifying the training dataset to balance bias, Li \etal \cite{li2018tell} suggested that network attention could be guided with biased data. However, self-guidance via soft mask application is ineffective if the region of interest (ROI) overlaps with the attention map.

In this paper, we incorporate user interaction to resolve co-occurrence issues when the attention map includes the ROI. The user directly specifies the region on which classifiers should focus, then the system re-trains the classifiers to focus on the specified region. We test this method using facial images; we find that the approach effectively addresses co-occurrence bias.

\section{Method}
\begin{figure}[!t]
\begin{center}
  \includegraphics[width=0.95\linewidth]{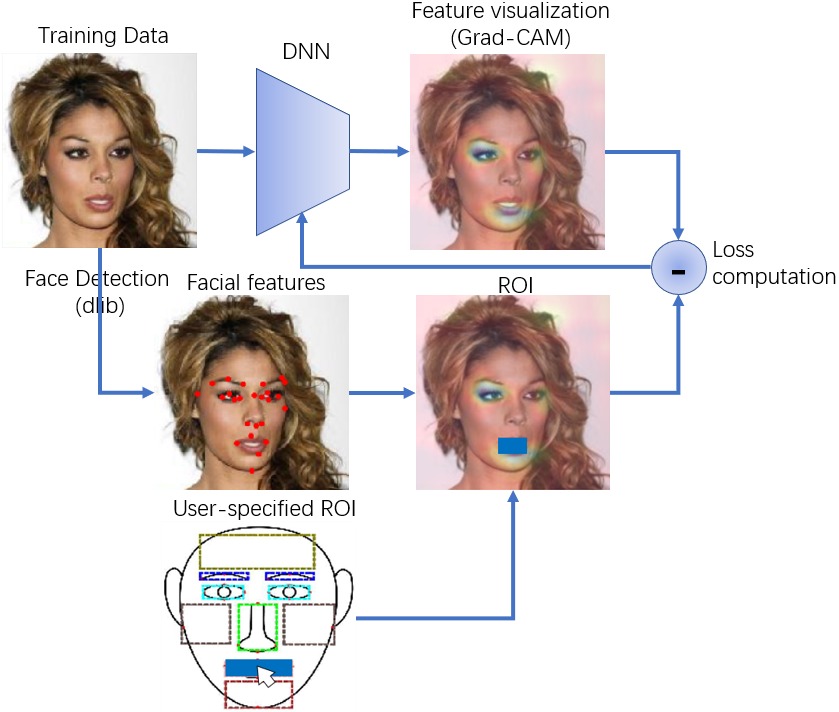}
\end{center}
   \caption{\bj{Overview of how we direct the attention of network on `Wearing Lipstick' images.}}
\label{fig:overview}
\end{figure}

An overview of our method is shown in the Figure \ref{fig:overview}. Given a pre-trained classification DNN, we visualize the activation in the model to localize the regions where network focuses on for some example images. If the network makes a classification based on biased features, the user manually specifies the correct region on a template. \bj{This will fine-tune the pre-trained network to focus on user's defined region and direct the attention of network accordingly.}

\textbf{Visualization of the feature maps.} 
\bj{Gradient-weighted Class Activation Mapping (Grad-CAM) \cite{selvaraju2017grad} is employed in our method, which uses class-specific gradient information to localize important regions in terms of classification.}

\begin{figure}[!t]
\begin{center}
  \includegraphics[width=0.85\linewidth]{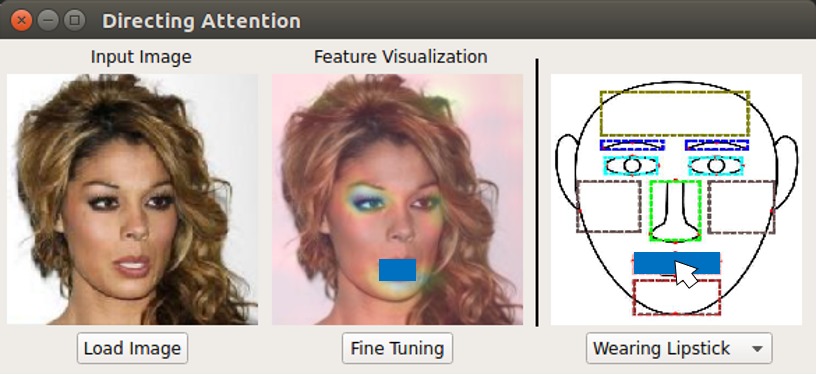}
\end{center}
   \caption{The user interface for the lipstick problem. After loading an input image and selecting a single class, the original image and the visualization are shown side-by-side. The user could select desired attention region(s) to fine-tune the pre-trained network.}
\label{fig:ui}
\end{figure}

\bj{\textbf{Specifying region of interest (ROI).}} 
\yx{We assume that one facial attribute corresponds to some facial regions. In order to conveniently specify the desired location that requires attention,}
we use Dlib\footnote{\url{http://dlib.net/}} to detect landmarks and segment the facial regions of each image. The user interface is shown in Figure \ref{fig:ui}. The user first identifies the most important region of an input image; this should include some of ten pre-defined rectangular facial regions. Then the entire specified rectangular region is used to calculate Grad-CAM loss. \bj{Note that}, the user only needs to select a rectangle in template face illustration.

\begin{figure}[h]
\begin{center}
  \includegraphics[width=0.85\linewidth]{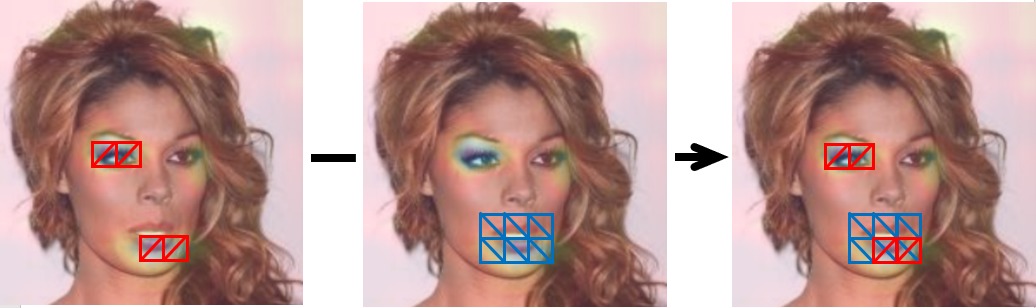}
\end{center}
   \caption{Grad-CAM loss in an image exemplifying the lipstick problem. Red boxes show the Grad-CAM regions and blue boxes indicate the specified regions.}
\label{fig:losses}
\end{figure}

\textbf{Loss function.} The pre-trained network is fine-tuned using a loss function, \bj{which} is a weighted combination of attribute loss and Grad-CAM loss, see Equation (\ref{equ:loss}). The attribute loss $loss_a$ is the difference between the combined binary cross entropy (BCE) of the predicted scores and the labels. The Grad-CAM loss $loss_g$ is computed by comparing the Grad-CAM and the user-specified regions. The positive parameters $(w_a, w_g)$ are \bj{balancing weights for} $loss_a$ and $loss_g$. Neurons with values $> 0.5$ on Grad-CAM visualization constitute the Grad-CAM set. The landmarks of the specified region are mapped onto grids that are the same size as the Grad-CAM layer. We use the Intersection over Union (IoU) loss concept \cite{yu2016unitbox} to evaluate the extent of $loss_g$; we calculate the ratio of the overlap areas yielded by prediction and ground truth (Figure \ref{fig:losses}). The prediction value is the Grad-CAM region $G$ value, and the ground truth is the specified facial region $S$, as shown in Equation (\ref{equ:overlap}). 

\begin{equation}\label{equ:loss}
  Loss = w_a \cdot loss_a + w_g \cdot loss_g
\end{equation}

\begin{equation}\label{equ:overlap}
  IoU_{loss} = - ln\left(\frac{G \bigcap S}{G \bigcup S}\right),
\end{equation}

\section{Experiments}

We tested our method using the CelebA dataset \cite{liu2015faceattributes}, a large-scale facial attributes dataset containing more than 200,000 celebrity images, each featuring 41 attribute annotations. We divided these images into three sets: training, validation, and testing set. To enhance the diversity, 
\bj{two image sets that showing mouth region only and concealing eyes with sunglasses are included.}
Let $a_p$ and $b_p$ denote two image sets positively annotated in terms of attributes $a$ and $b$, and let $a^c_p$ and $b^c_p$ denote two negative sets. Then the image sets $E_1 = a_p \bigcap b_p$ and $E_2 = a_p \bigcap b^c_p$ of the testing set are extracted to evaluate network fine-tuning. We used AlexNet for facial attribute classification task which contains five convolutional layers and three fully connected layers. The last conv-layer delivers the Grad-CAM results.

\begin{figure}[h]
\begin{center}
  \includegraphics[width=0.15\linewidth]{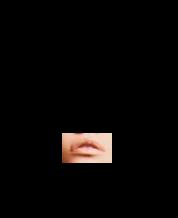}
  \includegraphics[width=0.15\linewidth]{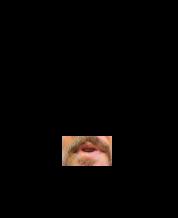}
  \includegraphics[width=0.15\linewidth]{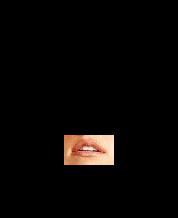}
  \includegraphics[width=0.15\linewidth]{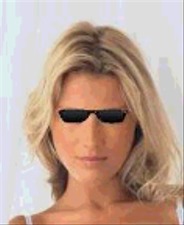}
  \includegraphics[width=0.15\linewidth]{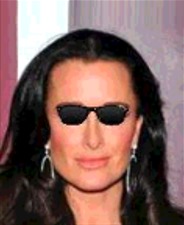}
  \includegraphics[width=0.15\linewidth]{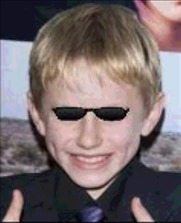}
\end{center}
   \caption{Examples of masked images and images with sunglasses.}
\label{fig:masked}
\end{figure}

We used three examples: `Heavy Makeup' \& `Wearing Lipstick', `High Cheekbones' \& `Smiling', and `Chubby' \& `Double Chin', in our experiments. For the attribute `Wearing Lipstick', the classification accuracy of the pre-trained network was about $92.90\%$ for the test set. Next, we specified that the network should focus only on the mouth; we fine-tuned the pre-trained AlexNet accordingly, accompanied by IoU loss. Accuracy improved to $93.25\%$. For the sunglasses set, the classification accuracy improved from $83.53\%$ to $86.61\%$. We also used two straightforward methods to modify the dataset in our comparison experiment. The network $N_o$ was trained using edited images of the mouth region only; the accuracy was about $56.88\%$. Then we mixed these images with the original training set to create a network that weighted mouth regions more highly; this trained network ($N_w$) \bj{reported} an accuracy of $93.14\%$.

\begin{figure*}[h]
\begin{center}
  Pre-trained ~~~~~~~~~~~~~~~~~~~~~~~~~~~~~~~~~~~~~~~~~~~~~~~~~~~~~~~~~~~~~~~~~~~~~~~~~~~~~ Fine-tuned \\
  \includegraphics[width=0.09\linewidth]{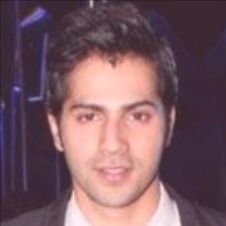}
  \includegraphics[width=0.09\linewidth]{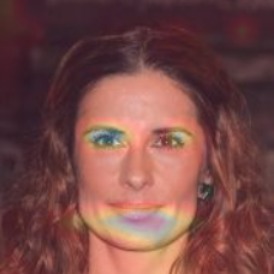}
  \includegraphics[width=0.09\linewidth]{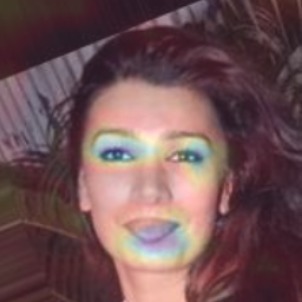}
  \includegraphics[width=0.09\linewidth]{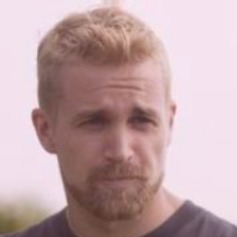}
  \includegraphics[width=0.09\linewidth]{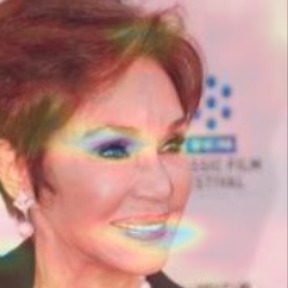}
  ~~~
  \includegraphics[width=0.09\linewidth]{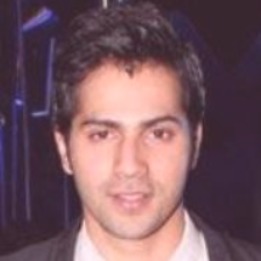}
  \includegraphics[width=0.09\linewidth]{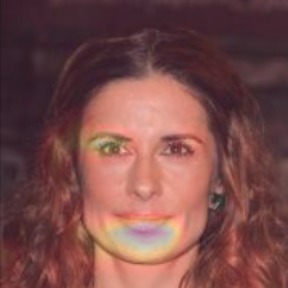}
  \includegraphics[width=0.09\linewidth]{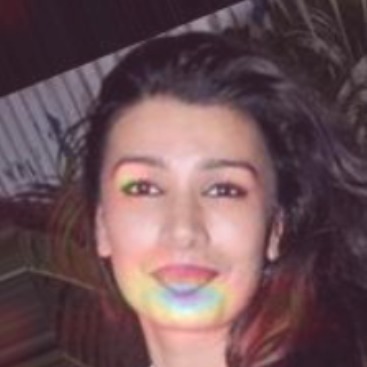}
  \includegraphics[width=0.09\linewidth]{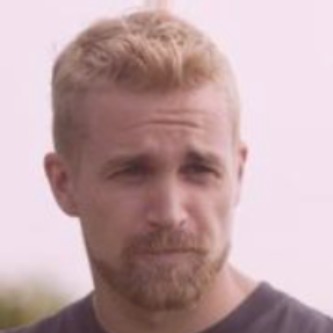}
  \includegraphics[width=0.09\linewidth]{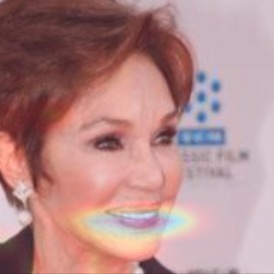}
  \\
  \includegraphics[width=0.09\linewidth]{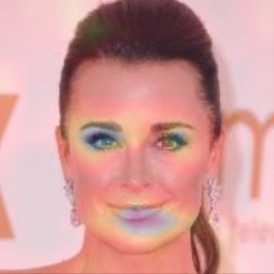}
  \includegraphics[width=0.09\linewidth]{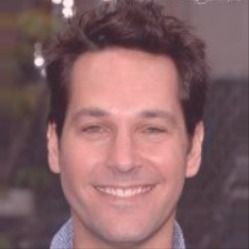}
  \includegraphics[width=0.09\linewidth]{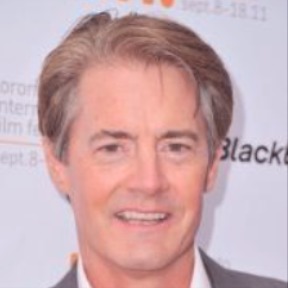}
  \includegraphics[width=0.09\linewidth]{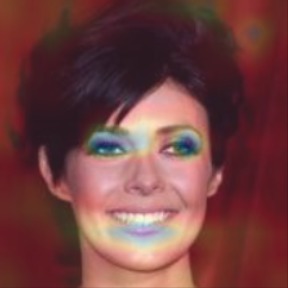}
  \includegraphics[width=0.09\linewidth]{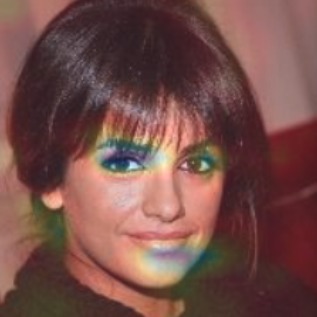}
  ~~~
  \includegraphics[width=0.09\linewidth]{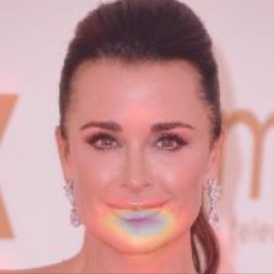}
  \includegraphics[width=0.09\linewidth]{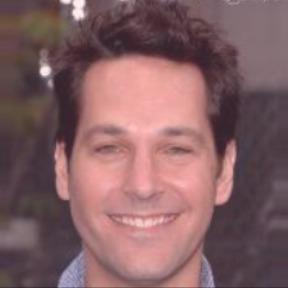}
  \includegraphics[width=0.09\linewidth]{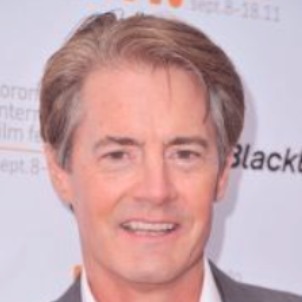}
  \includegraphics[width=0.09\linewidth]{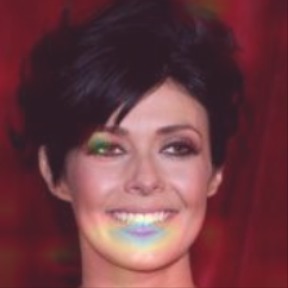}
  \includegraphics[width=0.09\linewidth]{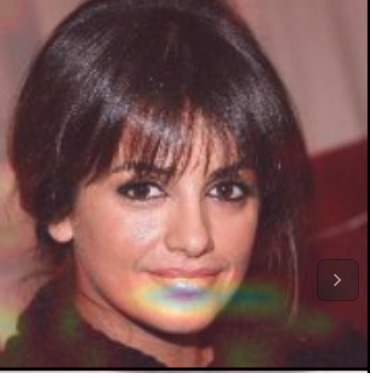}
  \\
  $N_o$ ~~~~~~~~~~~~~~~~~~~~~~~~~~~~~~~~~~~~~~~~~~~~~~~~~~~~~~~~~~~~~~~~~~~~~~~~~~~~~~~~~~~~~~~~~~~~~ $N_w$ \\
  \includegraphics[width=0.09\linewidth]{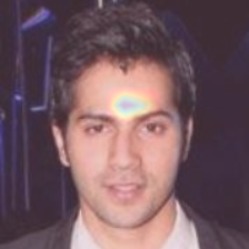}
  \includegraphics[width=0.09\linewidth]{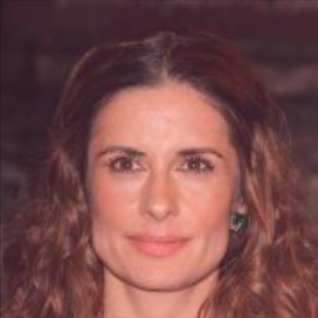}
  \includegraphics[width=0.09\linewidth]{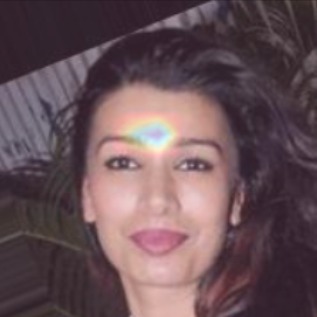}
  \includegraphics[width=0.09\linewidth]{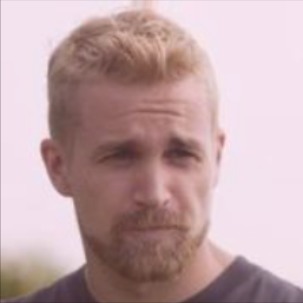}
  \includegraphics[width=0.09\linewidth]{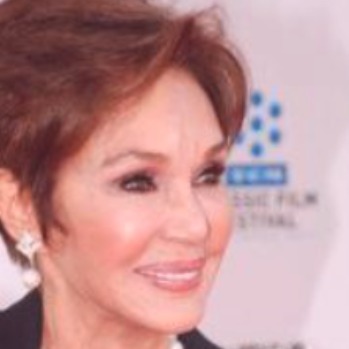}
  ~~~
  \includegraphics[width=0.09\linewidth]{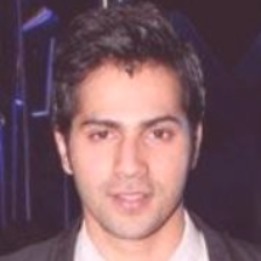}
  \includegraphics[width=0.09\linewidth]{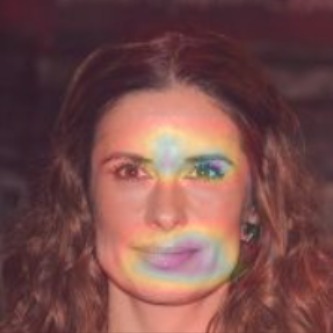}
  \includegraphics[width=0.09\linewidth]{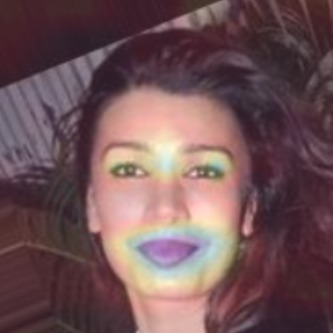}
  \includegraphics[width=0.09\linewidth]{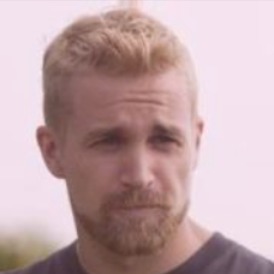}
  \includegraphics[width=0.09\linewidth]{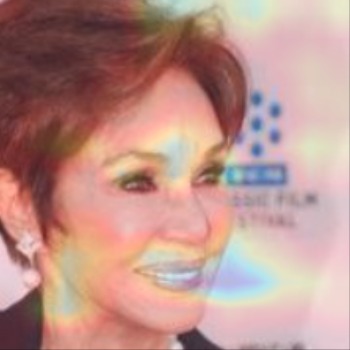}
  \\
  \includegraphics[width=0.09\linewidth]{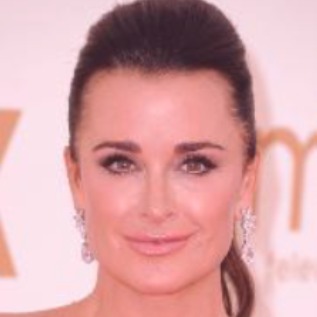}
  \includegraphics[width=0.09\linewidth]{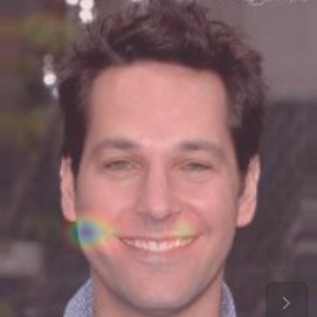}
  \includegraphics[width=0.09\linewidth]{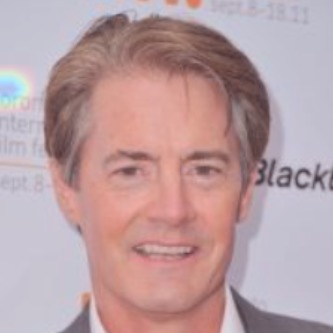}
  \includegraphics[width=0.09\linewidth]{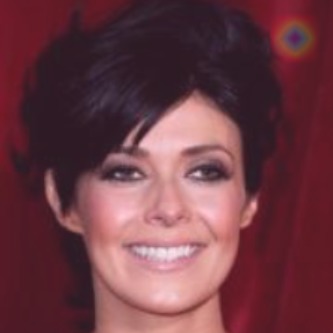}
  \includegraphics[width=0.09\linewidth]{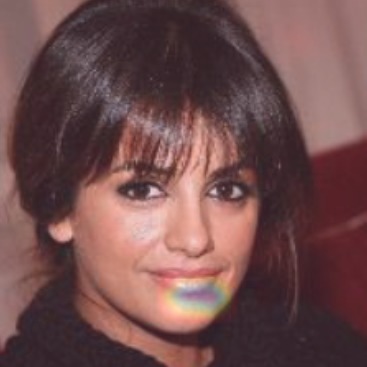}
  ~~~
  \includegraphics[width=0.09\linewidth]{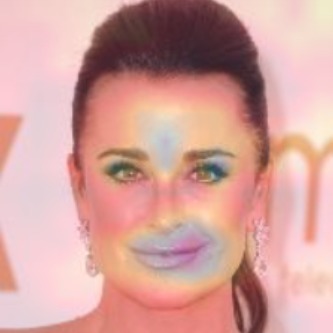}
  \includegraphics[width=0.09\linewidth]{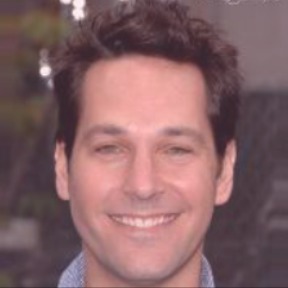}
  \includegraphics[width=0.09\linewidth]{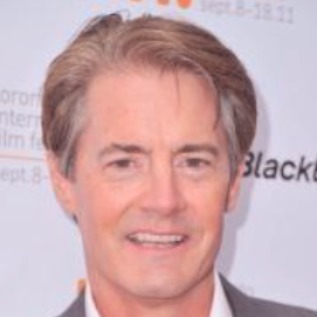}
  \includegraphics[width=0.09\linewidth]{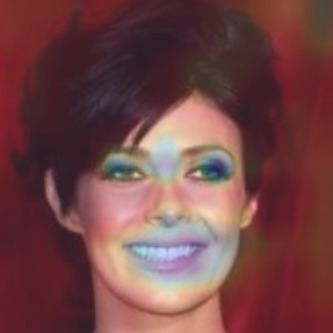}
  \includegraphics[width=0.09\linewidth]{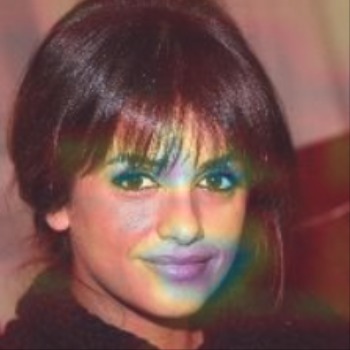}
\end{center}
   \caption{A comparison of results afforded by the four networks.}
\label{fig:compare}
\end{figure*}

The Grad-CAM results for 10 selected images of the test sets for the four networks are shown in Figure \ref{fig:compare}. It is clear that the pre-trained network classified `Wearing Lipstick' based not only on the mouth region but also by reference to the eyebrows and eyes. Using our method, fine-tuning significantly reduced dependence on eyebrows and eyes, emphasizing the mouth. The $N_o$ focused on the mouth only, but the accuracy of the test set was low. The $N_w$ was very accurate, but was affected by co-occurrence bias.

\begin{figure*}[ht]
\begin{center}
  $E_1$ ~~~~~~~~~~~~~~~~~~~~~~~~~~~~~~~~~~~~~~~~~~~~~~~~~~~~~~~~~~~~~~~~~~~~~~~~~~~~~~~~~~~~~~~~~~~~~ $E_2$ \\
  \includegraphics[width=0.09\linewidth]{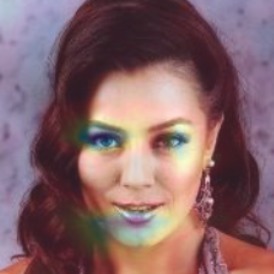}
  \includegraphics[width=0.09\linewidth]{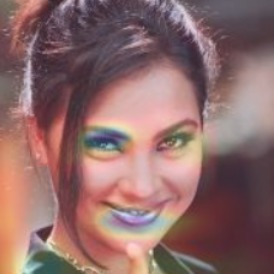}
  \includegraphics[width=0.09\linewidth]{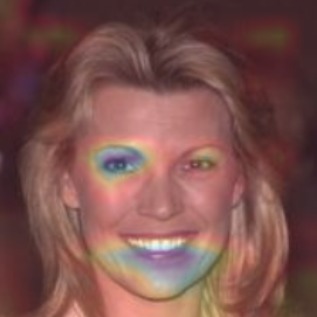}
  \includegraphics[width=0.09\linewidth]{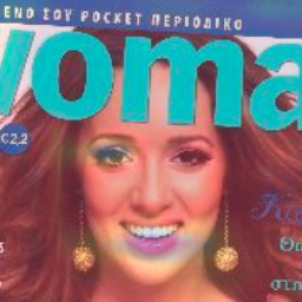}
  \includegraphics[width=0.09\linewidth]{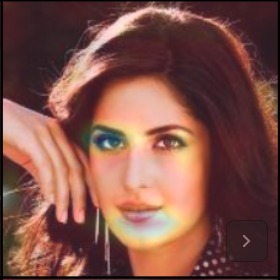}
  ~~~
  \includegraphics[width=0.09\linewidth]{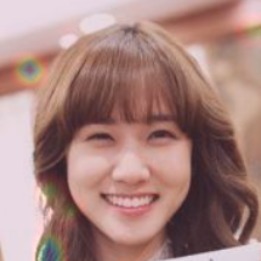}
  \includegraphics[width=0.09\linewidth]{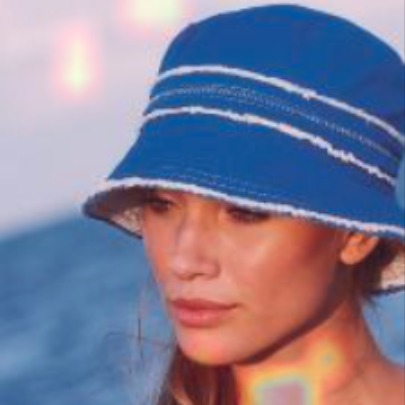}
  \includegraphics[width=0.09\linewidth]{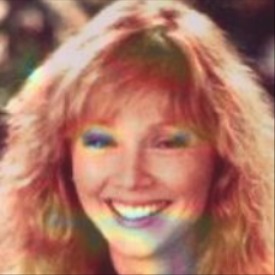}
  \includegraphics[width=0.09\linewidth]{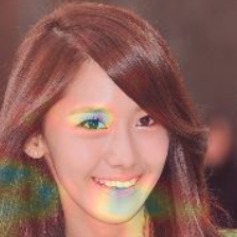}
  \includegraphics[width=0.09\linewidth]{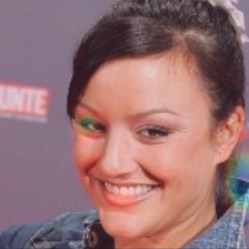}
  \\
  \includegraphics[width=0.09\linewidth]{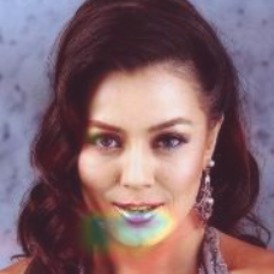}
  \includegraphics[width=0.09\linewidth]{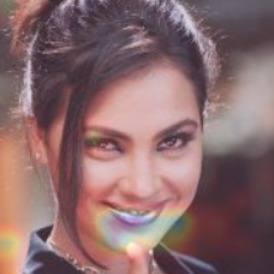}
  \includegraphics[width=0.09\linewidth]{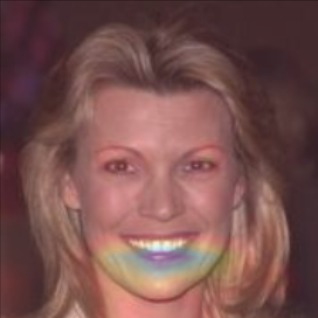}
  \includegraphics[width=0.09\linewidth]{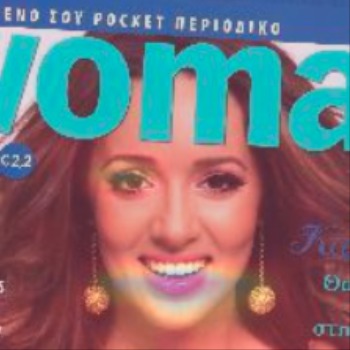}
  \includegraphics[width=0.09\linewidth]{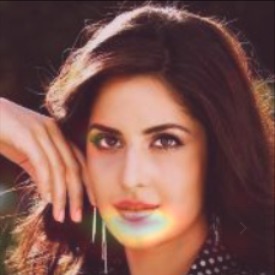}
  ~~~
  \includegraphics[width=0.09\linewidth]{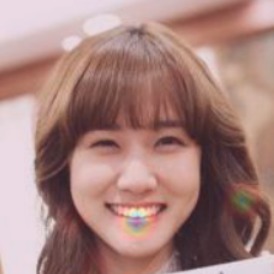}
  \includegraphics[width=0.09\linewidth]{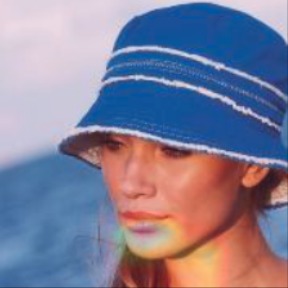}
  \includegraphics[width=0.09\linewidth]{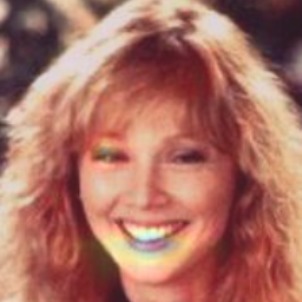}
  \includegraphics[width=0.09\linewidth]{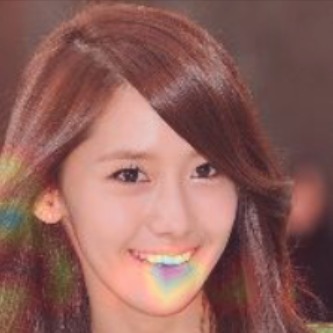}
  \includegraphics[width=0.09\linewidth]{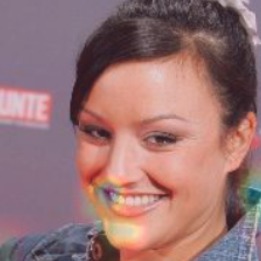}
  \\
  ~
  \\
  \includegraphics[width=0.09\linewidth]{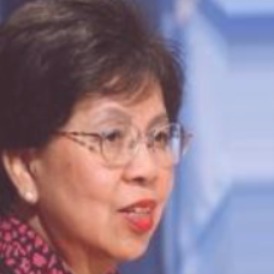}
  \includegraphics[width=0.09\linewidth]{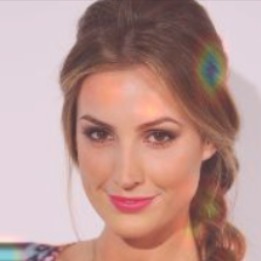}
  \includegraphics[width=0.09\linewidth]{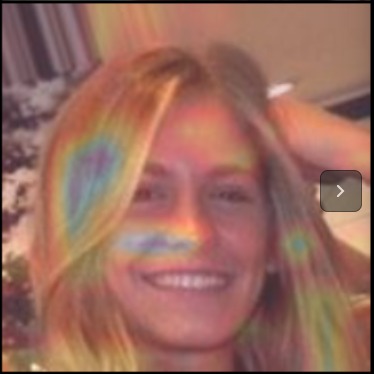}
  \includegraphics[width=0.09\linewidth]{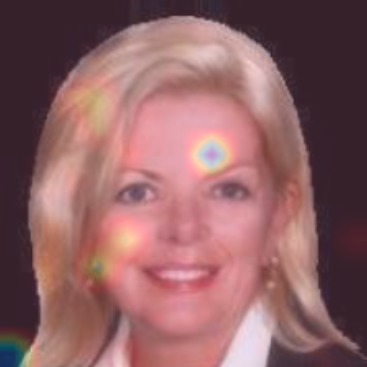}
  \includegraphics[width=0.09\linewidth]{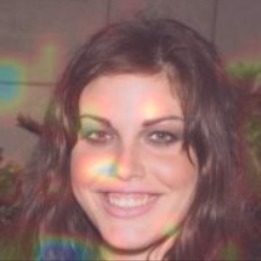}
  ~~~
  \includegraphics[width=0.09\linewidth]{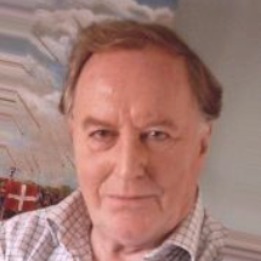}
  \includegraphics[width=0.09\linewidth]{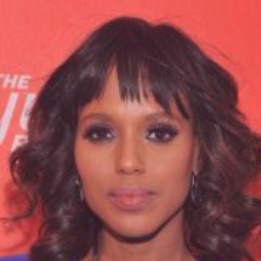}
  \includegraphics[width=0.09\linewidth]{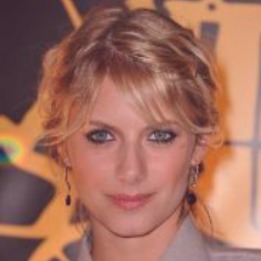}
  \includegraphics[width=0.09\linewidth]{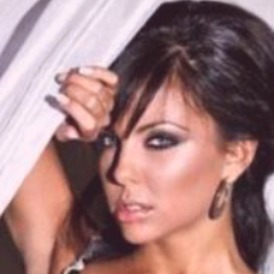}
  \includegraphics[width=0.09\linewidth]{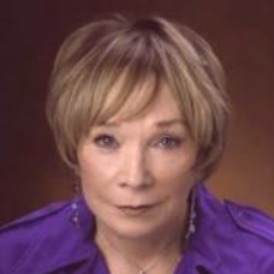}
  \\
  \includegraphics[width=0.09\linewidth]{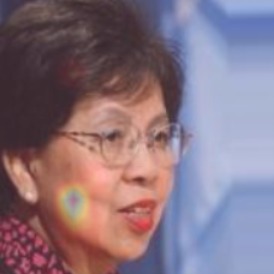}
  \includegraphics[width=0.09\linewidth]{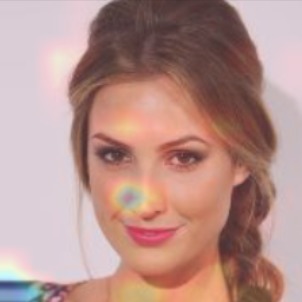}
  \includegraphics[width=0.09\linewidth]{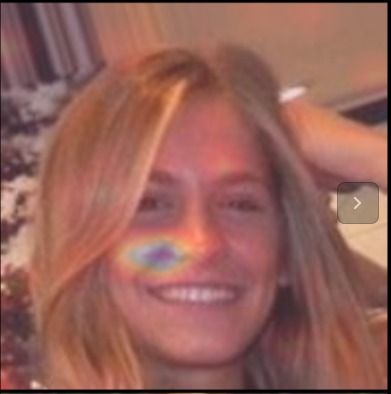}
  \includegraphics[width=0.09\linewidth]{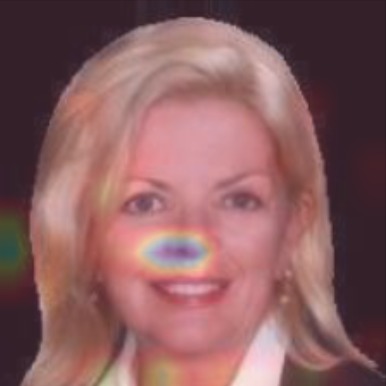}
  \includegraphics[width=0.09\linewidth]{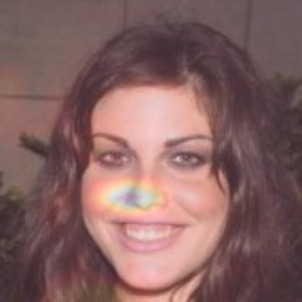}
  ~~~
  \includegraphics[width=0.09\linewidth]{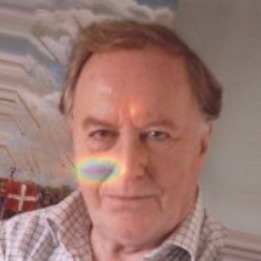}
  \includegraphics[width=0.09\linewidth]{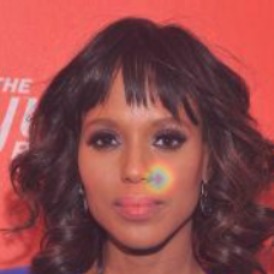}
  \includegraphics[width=0.09\linewidth]{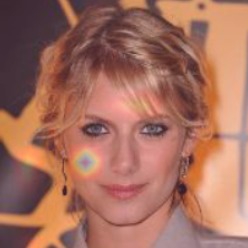}
  \includegraphics[width=0.09\linewidth]{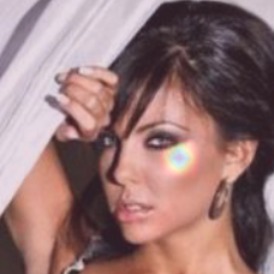}
  \includegraphics[width=0.09\linewidth]{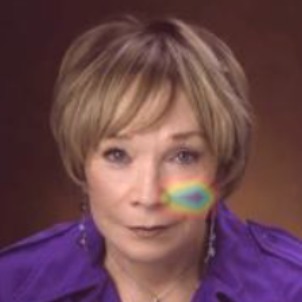}
  \\
  ~
  \\
  \includegraphics[width=0.09\linewidth]{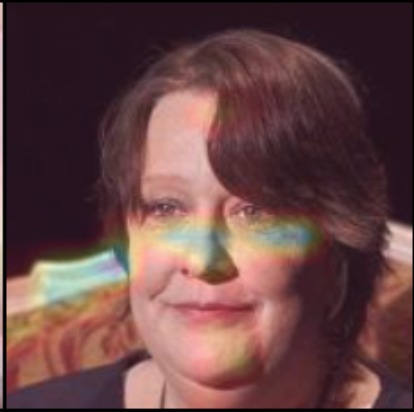}
  \includegraphics[width=0.09\linewidth]{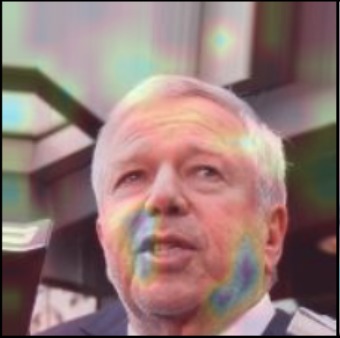}
  \includegraphics[width=0.09\linewidth]{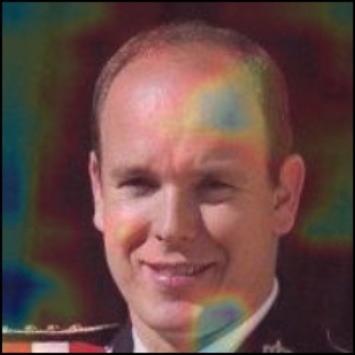}
  \includegraphics[width=0.09\linewidth]{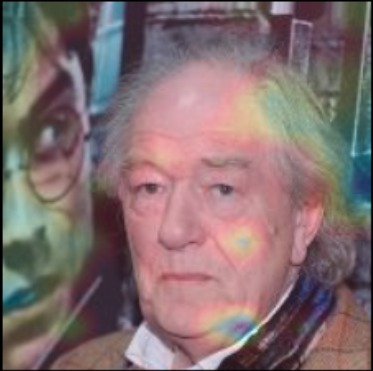}
  \includegraphics[width=0.09\linewidth]{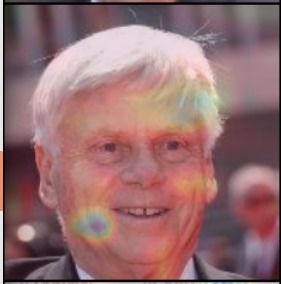}
  ~~~
  \includegraphics[width=0.09\linewidth]{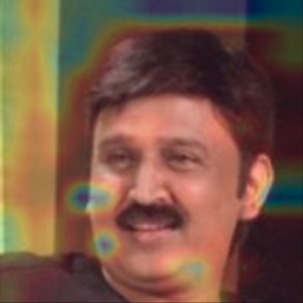}
  \includegraphics[width=0.09\linewidth]{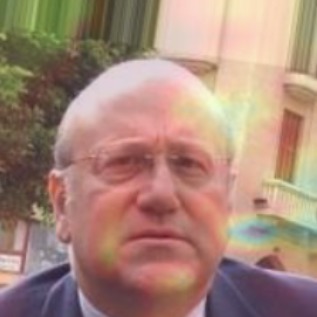}
  \includegraphics[width=0.09\linewidth]{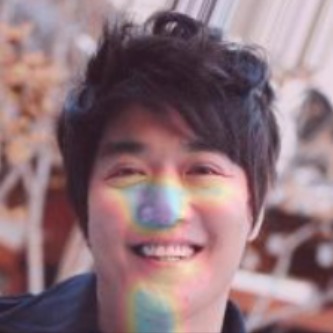}
  \includegraphics[width=0.09\linewidth]{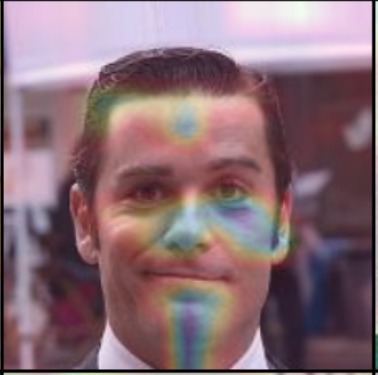}
  \includegraphics[width=0.09\linewidth]{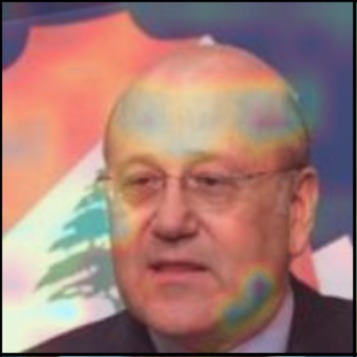}
  \\
  \includegraphics[width=0.09\linewidth]{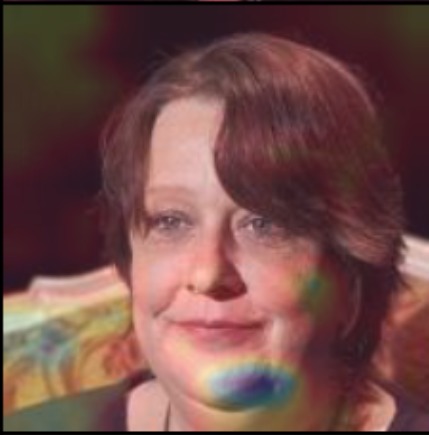}
  \includegraphics[width=0.09\linewidth]{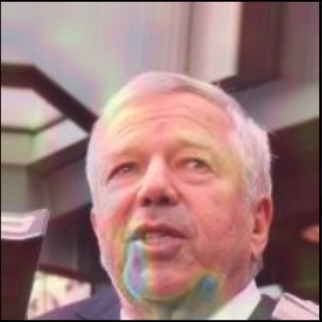}
  \includegraphics[width=0.09\linewidth]{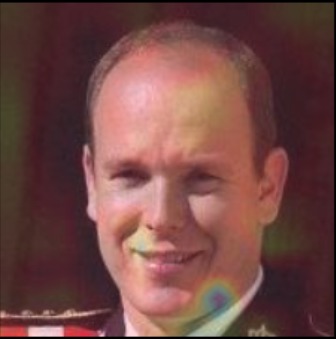}
  \includegraphics[width=0.09\linewidth]{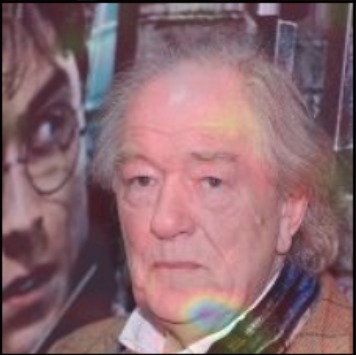}
  \includegraphics[width=0.09\linewidth]{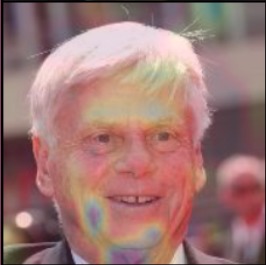}
  ~~~
  \includegraphics[width=0.09\linewidth]{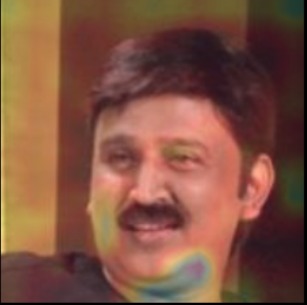}
  \includegraphics[width=0.09\linewidth]{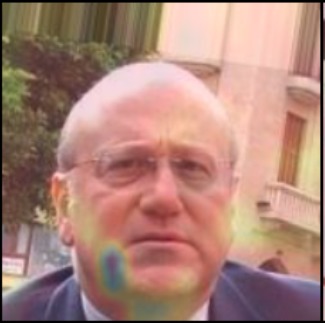}
  \includegraphics[width=0.09\linewidth]{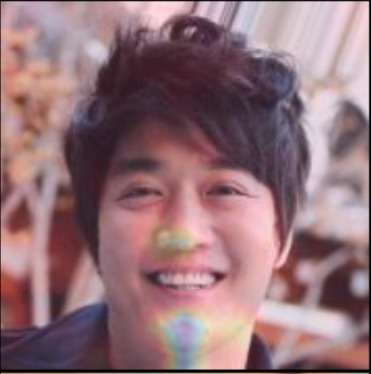}
  \includegraphics[width=0.09\linewidth]{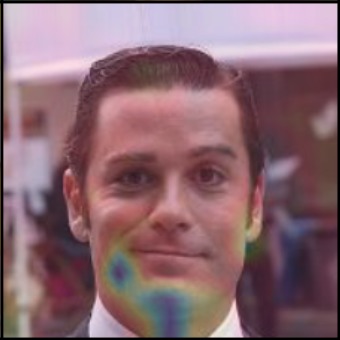}
  \includegraphics[width=0.09\linewidth]{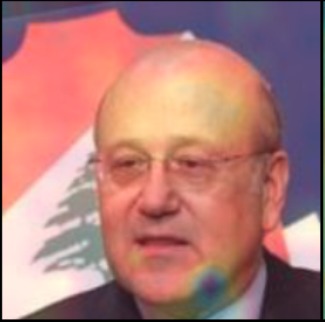}
\end{center}
   \caption{Comparison of the Grad-CAM results derived using the two image sets $E_1$ and $E_2$. The attributes of `Wearing Lipstick', `High Cheekbones', and `Double Chin’ are shown in the first to third blocks, respectively. In each block, the results afforded by the pre-trained network are shown in the first row, and those from the fine-tuned network are given in the second row.}
\label{fig:result}
\end{figure*}

We compared the accuracies of the pre-trained and fine-tuned networks using the image sets $E_1$ and $E_2$ shown in Table \ref{tab:accuracy}. The Grad-CAM results for five images from each set are shown in Figure \ref{fig:result}. Images featuring the attributes of both `Wearing Lipstick' and `Heavy Makeup' (from $E_1$) clearly differ in terms of DNN measured attention before and after fine-tuning. For `Wearing Lipstick' but without `Heavy Makeup' images (from $E_2$), the fine-tuned (but not the pre-trained) network detected `lipstick'. Thus, the former network exhibited better transferability. The experimental results for the attributes `High Cheekbones' \& `Smiling' and `Double Chin' \& `Chubby' are shown in Figure \ref{fig:result}.

\begin{table}[h]
\caption{The accuracies afforded by each network (three examples).}
\label{tab:accuracy}
\centering
\begin{tabular}{cccc}
\multicolumn{4}{c}{`Wearing Lipstick' \& `Heavy Makeup'} \\
\hline
 Network & test set & $E_1$ & $E_2$ \\
\hline
Pre-trained & 92.90\% & 98.06\% & 82.17\% \\
Center loss & 93.26\% & 98.29\% & 83.23\% \\
IoU loss & 93.25\% & 98.31\% & 83.31\% \\
\hline
\\
\multicolumn{4}{c}{`High Cheekbones' \& `Smiling'} \\
\hline
Network & test set & $E_1$ & $E_2$ \\
\hline
Pre-trained & 63.53\% & 70.22\% & 47.41\% \\
IoU loss & 65.56\% & 71.42\% & 61.33\% \\
\hline
\\
\multicolumn{4}{c}{`Double Chin' \& `Chubby'} \\
\hline
Network & test set & $E_1$ & $E_2$ \\
\hline
Pre-trained & 84.18\% & 90.92\% & 81.77\% \\
IoU loss & 84.79\% & 94.32\% & 86.45\% \\
\hline
\end{tabular}
\end{table}

\subsection{Training and test details}
We performed all experiments using PyTorch running on a PC featuring a GPU GTX 1080 processor. We trained AlexNet via early stopping of the training and validation sets. Optimization was achieved using the stochastic gradient descent (SGD) (with momentum) method. The learning rate and the momentum were 0.01 and 0.9, respectively, during both training and fine-tuning. The batch size was 256. All fine-tuning results were derived by performing single-epoch AlexNet runs. The training and fine-tuning epoch times were about 6 and 10 min, respectively. The mouth region weight was triple that of other facial parts. We set $(w_a, w_g)$ to $(1,0,5.0)$, $(1.0, 4.0)$, and $(1.0, 3.0)$ in the `Wearing Lipstick', `High Cheekbones', and `Double Chin' experiments, respectively.

\section{Conclusion}
We developed a method whereby a user can manually instruct a pre-trained DNN  network to focus on only relevant features; this eliminates co-occurrence bias that may be present in a dataset. We used the CelebA dataset to solve the lipstick problem; our method reduced the effects of co-occurring features in the pre-trained network; classification accuracy improved greatly. However, face landmark recognition accuracy affected our results. In the future, we will eschew landmarks. Additionally, we will test our method employing other DNN models (VGG and ResNet).

\section{Acknowledgement}
This work was supported by JST CREST Grant Number JPMJCR17A1, Japan.

{\small
\bibliographystyle{ieee_fullname}
\bibliography{egbib}
}

\end{document}